\pdfoutput=1
\documentclass[10pt,twocolumn,letterpaper]{article}

\usepackage{cvpr}              

\usepackage{graphicx}
\usepackage{amsmath}
\usepackage{amssymb}
\usepackage{booktabs}

\usepackage{cvpr}
\usepackage{blindtext}
\usepackage{caption}

\usepackage[pagebackref,breaklinks,colorlinks]{hyperref}

\usepackage{bm}
\usepackage{multirow}

\usepackage[capitalize]{cleveref}
\crefname{section}{Sec.}{Secs.}
\Crefname{section}{Section}{Sections}
\Crefname{table}{Table}{Tables}
\crefname{table}{Tab.}{Tabs.}

\def\cvprPaperID{3750} 
\def\confName{CVPR}
\def\confYear{2024}

\begin{document}
\title{TrainerAgent: Customizable and Efficient Model Training \\ through LLM-Powered Multi-Agent System}
\author{
Haoyuan Li\textsuperscript{\rm 1}, 
Hao Jiang\textsuperscript{\rm 1}\thanks{Corresponding author.}, 
Tianke Zhang\textsuperscript{\rm 1,2}, 
Zhelun Yu\textsuperscript{\rm 1}, 
Aoxiong Yin\textsuperscript{\rm 3}, \\
Hao Cheng\textsuperscript{\rm 1}, 
Siming Fu\textsuperscript{\rm 1}, 
Yuhao Zhang\textsuperscript{\rm 1}, 
Wanggui He\textsuperscript{\rm 1} \\
\textsuperscript{\rm 1}Taotian Group, \textsuperscript{\rm 2}Tsinghua University, \textsuperscript{\rm 3}Zhejiang University}





\maketitle




\begin{abstract}

Training AI models has always been challenging, especially when there is a need for custom models to provide personalized services. Algorithm engineers often face a lengthy process to iteratively develop models tailored to specific business requirements, making it even more difficult for non-experts. The quest for high-quality and efficient model development, along with the emergence of Large Language Model (LLM) Agents, has become a key focus in the industry. Leveraging the powerful analytical, planning, and decision-making capabilities of LLM, we propose a TrainerAgent system comprising a multi-agent framework including Task, Data, Model and Server agents. These agents analyze user-defined tasks, input data, and requirements (e.g., accuracy, speed), optimizing them comprehensively from both data and model perspectives to obtain satisfactory models, and finally deploy these models as online service. 
Experimental evaluations on classical discriminative and generative tasks in computer vision and natural language processing domains demonstrate that our system consistently produces models that meet the desired criteria. Furthermore, the system exhibits the ability to critically identify and reject unattainable tasks, such as fantastical scenarios or unethical requests, ensuring robustness and safety. 
This research presents a significant advancement in achieving desired models with increased efficiency and quality as compared to traditional model development, facilitated by the integration of LLM-powered analysis, decision-making, and execution capabilities, as well as the collaboration among four agents.
We anticipate that our work will contribute to the advancement of research on TrainerAgent in both academic and industry communities, potentially establishing it as a new paradigm for model development in the field of AI. 

\end{abstract}

\section{Introduction}

The rapid advancement of artificial intelligence (AI) has revolutionized numerous industries, enabling personalized and efficient services that were once unimaginable. However, the process of training AI models to meet specific business requirements remains a daunting and time-consuming challenge. This is particularly pertinent for non-experts who struggle to navigate the intricacies of model development and customization. Bridging this gap between user needs and model development has become a pressing concern in the AI industry. 

Nowadays, autonomous agents \cite{park2023generative, zhuge2023mindstorms,cai2023large,wang2023unleashing,li2023camel,du2023improving,liang2023encouraging,hao2023chatllm} utilizing Large Language Models (LLMs) offer promising opportunities to enhance and replicate human workflows, which seems to able to solve ease the concern above. 
Specially, HuggingGPT \cite{shen2023hugginggpt}, a framework that employs large language models like ChatGPT as controllers to integrate various specialized AI models for complex tasks. It uses natural language as an interface to streamline task execution across different domains and modalities, demonstrating the potential for more advanced AI systems. MetaGPT \cite{hong2023metagpt} introduces a meta-programming framework that enhances LLM-based multi-agent systems by incorporating standardized workflows to reduce logic errors and increase task efficiency. It achieves superior performance by assigning specialized roles to agents for collaborative problem-solving, outperforming existing chat-based solutions in complex benchmarks. 
AutoGen \cite{wu2023autogen} provides an open-source platform for building complex LLM applications, allowing for inter-agent communication and a blend of LLM capabilities, human inputs, and additional tools. It enables the customization of conversational patterns and agent behaviors, demonstrating its versatility and effectiveness across a wide range of fields, from technical domains to creative industries. 
However, the current agent system is unable to satisfactorily accomplish the construction of specific requirements, from user needs to model training and deployment, particularly in terms of model training. It lacks dedicated mechanisms to ensure the success rate of system operation and the training effectiveness of the final model. 
Although there are also some works that specialize in training models using LLMs, they still have significant limitations. AutoML-GPT \cite{zhang2023automlgpt} merges the power of LLM with expert system insights to automate AI model training, encompassing data processing to design and experiment execution. It simplifies the development of AI solutions by using standardized prompts based on comprehensive model and data descriptions. This unified approach has proven effective across various AI tasks, including those in language and vision, and excels in adapting to and tuning for new datasets as evidenced by rigorous testing. However, it requires fixed model inputs, which is rigid, demanding a high understanding of algorithms for users, while our system accepts natural language inputs, automatically comprehends the specific AI models involved, and performs training and optimization. Prompt2Model \cite{viswanathan2023prompt2model} advances the field by proposing a method that uses natural language task descriptions to train specialized models, offering competence with fewer computational resources than LLM. It retrieves existing datasets, generates additional data using LLMs, and fine-tunes models for improved performance. However, Prompt2Model has limitations in scalability, lack of consideration for user private databases, and reliance on huggingface. It is also limited to NLP tasks and lacks flexibility. 

\begin{figure*}[t]
\begin{center}
\includegraphics[width=2\columnwidth]{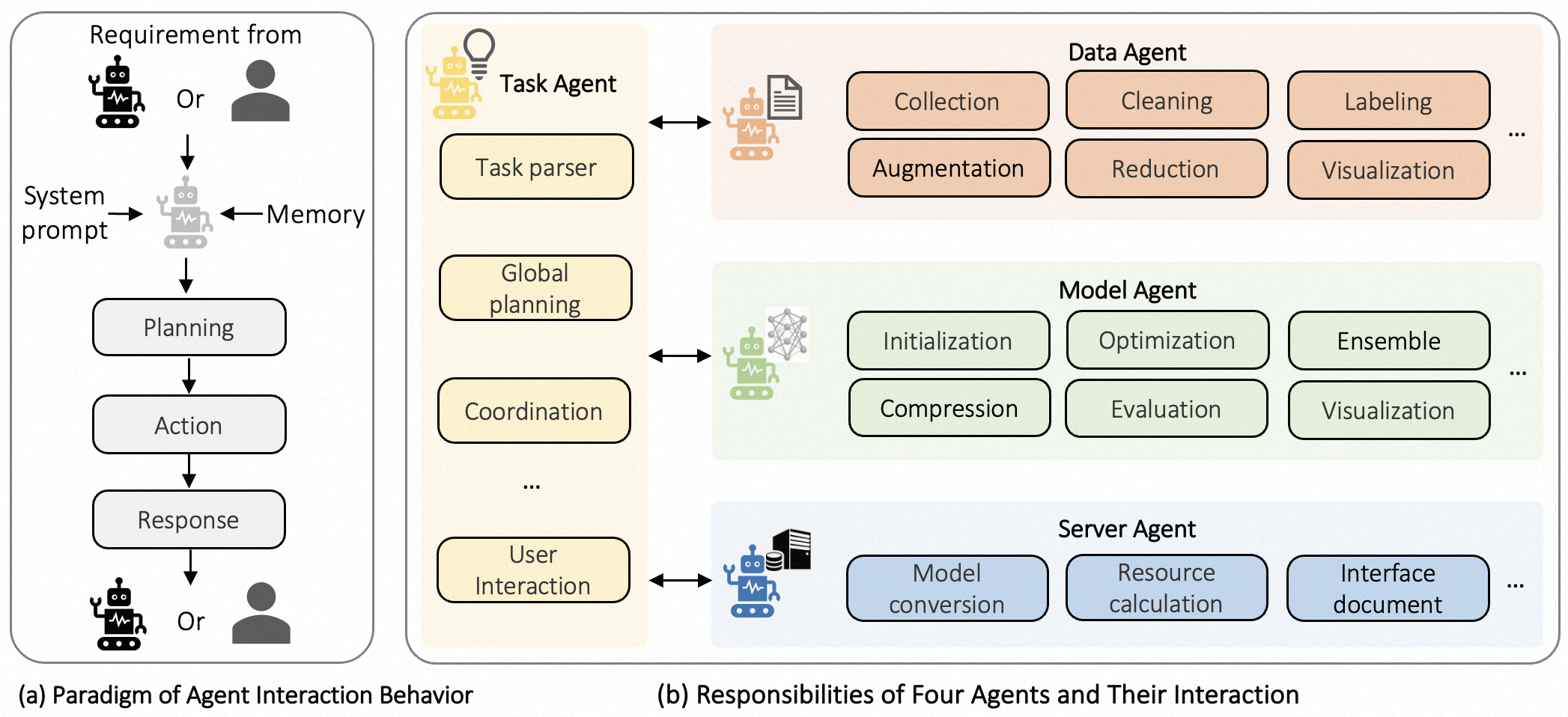}
\end{center}
   \caption{Interaction and Responsibilities of Agents.}
\label{fig:f2}
\end{figure*}

To build an intelligent system that can directly comprehend user-customized requirements and efficiently accomplish model training and deployment with enhanced flexibility, we propose TrainerAgent, a cutting-edge, customizable, and highly efficient model training system powered by groundbreaking LLM-powered Agents. 
Leveraging the remarkable analytical, scheduling, and decision-making capabilities of LLM, our system aims to revolutionize the way models are developed and deployed. By introducing a multi-agent framework comprising Task, Data, Model, and Server agents, TrainerAgent offers a comprehensive solution that optimizes models from both data and model perspectives, resulting in highly satisfactory outcomes. Specifically, The Task Agent acts as a hub, coordinating the activities of the other agents and interacting with the user, responsible for task parsing, global planning, coordination among agents, and user interaction. It parses user-defined tasks, develops a comprehensive plan for model development, coordinates agent activities, and provides a user-friendly interface. The Data Agent handles various data processing operations such as collection, cleaning, labeling, augmentation, reduction, and visualization. It works in collaboration with the Task Agent, receiving data processing requirements and instructions, and autonomously planning and executing these operations. The Model Agent is responsible for model initialization, optimization, ensemble, compression, evaluation, and visualization. It selects appropriate pre-trained models, optimizes their performance, conducts model compression, evaluates their performance, and provides visual representations and summaries of the models. The Server Agent handles model deployment based on user-defined online service requirements. It estimates resource needs, performs model conversion for compatibility and efficiency, and prepares interface documents for seamless integration with various applications and systems. 
And each agent is composed of several components and is provided with a system prompt and Standard Operating Procedures (SOPs) to guide their actions. The agents analyze requirements, plan their actions, and autonomously complete complex subtasks as Figure \ref{fig:f2} shown.

To evaluate the effectiveness of TrainerAgent, we conducted rigorous experimental evaluations on classical discriminative and generative tasks within the domains of computer vision (CV) and natural language processing (NLP) as Figure \ref{fig:vg} and \ref{fig:text} shown.  The results consistently demonstrated that our system produces exceptional models that meet the desired criteria. 
The qualitative analysis of the Visual Grounding, Image Generation and Text Classification task in the proposed TrainerAgent system demonstrates its ability to effectively handle internally constructed tasks, perform preliminary planning, and facilitate collaboration among different agents. The specialized agents also showcase their competence in fulfilling their assigned responsibilities. These features collectively contribute to the overall functionality and effectiveness of the TrainerAgent system.
Moreover, TrainerAgent showcased its remarkable ability to identify and reject unattainable tasks, ensuring the robustness and safety of the model development. 

Our research makes several significant contributions to the field of AI model development. Firstly, we introduce a novel system that automates the entire process, from requirement analysis to model training and deployment. This is the first of its kind and addresses the challenges faced by algorithm engineers in developing custom models for personalized services. Secondly, our approach utilizes a multi-agent framework comprising Task, Data, Model, and Server agents. These agents work collaboratively, each with their specific roles, to optimize user-defined tasks, input data, and requirements. This comprehensive optimization, considering both data and model perspectives, ensures the generation of satisfactory models that meet desired criteria such as accuracy and speed. Lastly, our system undergoes extensive experimental evaluations in computer vision and natural language processing domains. These evaluations demonstrate the consistent production of high-quality models that meet the desired criteria. Additionally, our system showcases the remarkable ability to identify and reject unattainable tasks, ensuring robustness and safety. We anticipate that our research will have a substantial impact on both academic and industry communities and establish the TrainerAgent system as a new paradigm for model development in AI.

\section{TrainerAgent} 
\subsection{Framework} 


In Section 1, as we have mentioned, our system can understand user's intent and ultimately train a model that satisfies the user's requirements based on four agents. 
Next, we will introduce how the entire system operates. 
Firstly, like most LLM-powered agents \cite{park2023generative, zhuge2023mindstorms,cai2023large,wang2023unleashing,li2023camel,du2023improving,liang2023encouraging,hao2023chatllm}, each agent in our system comprises the following components: profile, memory, perception, planning, action, and response, as illustrated in Figure \ref{fig:f2}(a).
Specifically, our agents are initially fed a system prompt as profile, informing them of the system overview and their responsibilities, and encoding Standard Operating Procedures (SOPs) \cite{belbin2012team,manifesto2001manifesto,demarco2013peopleware} into prompts . 
Moreover, during the interaction of Agents, the current requirements from user or other agents, as well as the memory of all past system interactions, are fed into the current agent. 
It then analyzes the current requirements, and enters the planning phase, organizing thoughts, set objectives, and determine the steps needed to achieve those objectives. Agents can also modify their plans through introspection to adapt to current circumstances. 
Next, the agent will take action based on the results of planning, and ultimately responds to the agent or user who provided the requirement. Through these operations, an agent can autonomously complete complex subtasks through various tools. 

However, the journey from business requirement identification to the final model deployment in the actual business scenario is not simple, involving numerous complex analysis and optimization. Based on our preliminary experiments, it is challenging and insufficientfor a single Agent to meet user requirements efficiently and effectively. 
Therefore, in our framework, we break down the entire process into four parts: task parsing and planning, data acquisition and analysis, model training and testing, and service deployment. These are implemented collaboratively by Task, Data, Model, and Server Agents respectively, as shown in Figure \ref{fig:f2}(b). 
Among them, the Task Agent acts as a hub, with all other agents interacting through it. It also interacts with the user, while the other three agents only focus on their specific tasks. 
Next, we will introduce the specific responsibilities for the four agents.

\subsection{Responsibility of Each Agent} 

{\noindent \bf Task Agent}

Task agent is the core agent in the TrainerAgent system, responsible for task parsing, global planning, coordination, and user interaction to ensure efficient and effective model development.
Firstly, the Task agent conducts task parsing, which involves parsing the user-defined tasks and extracting relevant information. This process includes identifying the specific goals and requirements of the tasks, such as the desired model accuracy, speed, or any other specific criteria. The parsed tasks are then transformed into a structured JSON format, enabling effective communication and collaboration with the other agents for further analysis and processing. 
Once the tasks are parsed, the Task agent engages in global planning. This step involves developing a comprehensive plan for model development that takes into account the parsed tasks, available input data, and the capabilities of the other agents. The Task agent assesses the feasibility and potential challenges associated with the tasks, considering factors such as data availability, computational resources, and model complexity. This planning phase aims to optimize the model development process and ensure that the subsequent steps are well-informed and aligned with the user's requirements.
Furthermore, the Task agent plays a pivotal role in coordinating the activities of the other agents within the system. It acts as a central coordinator, orchestrating the collaboration and communication between the Data, Model, and Server agents. This coordination ensures that the tasks are processed efficiently, and the agents work in tandem towards achieving the desired models. The Task agent schedules and assigns tasks to the relevant agents, monitors their progress, and resolves any conflicts or dependencies that may arise.
In addition to its coordination role, the Task agent also facilitates user interaction. It provides a user-friendly interface that allows users to interact with the TrainerAgent system. Users can provide feedback, refine their requirements, or monitor the progress of model development through this interface. 


{\noindent \bf Data Agent}

The Data agent plays a crucial role in the TrainerAgent system, primarily responsible for processing various types of data. To facilitate effective data processing, we have developed an extensive internal knowledge base within the Data Agent. This knowledge base encompasses a wide range of data modalities, including tabular, image, text, audio, and video data. It equips the agent with the understanding of which tools and techniques to employ for different types of data and specific processing scenarios. In cases where a suitable operation is not readily available in the knowledge base, the Data agent conducts online searches to find appropriate approaches. 
The Data agent operates in collaboration with the Task agent, receiving data processing requirements and instructions from the Task agent. Based on these requirements, the Data agent autonomously performs planning and action to execute the necessary operations. 
Specifically, the Data agent is responsible for data collection, which involves gathering relevant data from various sources such as internal databases or web scraping. This ensures a diverse and comprehensive dataset for model development. 
Furthermore, the Data agent conducts data cleaning, which focuses on removing noise, outliers, and inconsistencies from the collected data as well as correcting the annotation. This step aims to enhance the quality and reliability of the dataset, ensuring that subsequent modeling processes are based on clean and accurate data. 
Moreover, on scenarios where annotated data is insufficient, the Data agent possesses the capability to perform automatic data labeling. For instance, the Data agent can employ methods based on pre-training large-scale models to generate preliminary labels for various types of data, enabling the model to learn from a larger and more diverse dataset. 
Additionally, the Data agent performs data augmentation, which involves generating additional training samples by applying various transformations and modifications to the existing data. This technique helps to increase the diversity and generalization capability of the dataset, leading to improved model performance. 
Also, the Data agent conducts data reduction, which focuses on reducing the dimensionality or size of the dataset while preserving its key information. This step is particularly useful when dealing with large datasets or computationally intensive models, allowing for more efficient model training. 
Lastly, the Data agent facilitates data visualization, providing visual representations and summaries of the dataset to aid in data exploration and understanding. This enables users to gain insights into the data distribution and patterns, assisting in making informed decisions throughout the model development process. 




\begin{figure*}[t]
\begin{center}
\includegraphics[width=2.1\columnwidth]{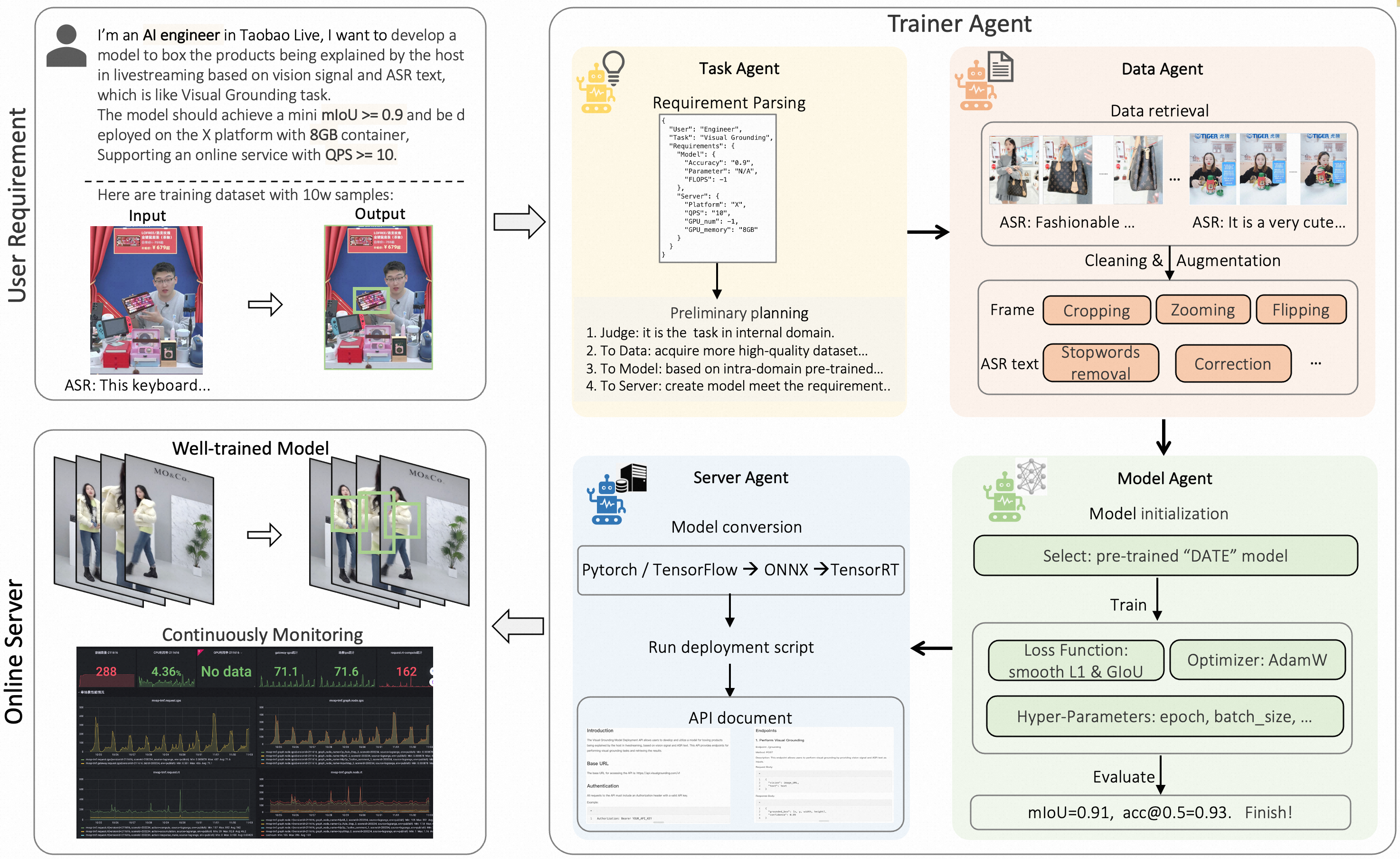}
\end{center}
   \caption{Qualitative Analysis of Visual Grounding Task. 
   The user presents a task to develop a model for Visual Grounding in live streaming, with specific performance and deployment requirements, and the Task Agent parses these requirements and initiates a preliminary planning. The Data Agent retrieves relevant Product Grounding dataset from internal databases and enhances it with image and text preprocessing techniques. The Model Agent then selects a pre-trained model from an internal library, trains and evaluates it against the set criteria. The Server Agent converts the model's format for deployment, estimates online resource required, sets up the service infrastructure on the specified platform, writes the API document, and establishes continuously monitoring mechanisms. The result is a well-trained model capable of providing an online service for product grounding in live streaming.}
\label{fig:vg}
\end{figure*}

{\noindent \bf Model Agent}

The Model agent is responsible for training and validating models. Similar to the Data agent, the Model agent receives task requirements and instructions from the Task agent. It autonomously performs planning and takes action based on these inputs.
Specifically, the Model agent is responsible for model initialization, which involves the selection of appropriate pre-trained models for specific tasks. The internal model repository including a comprehensive collection of pre-trained models suitable for different tasks and the huggingface model retriever provide a vast array of pre-trained models, allowing the Model agent to identify the most suitable ones based on the task requirements. 
Furthermore, the Model agent carries out optimization processes to enhance the performance of the selected models, along with standardized training scripts based on huggingface.  Leveraging the internal training knowledge base we built, the Model agent automates various optimization techniques such as hyperparameter tuning, learning rate scheduling, and regularization. This ensures that the models are trained effectively and efficiently. The Model agent can leverage ensemble methods to improve model performance if needed. 
Moreover, the Model agent performs model compression, aiming to reduce the size and complexity of the models without significant performance degradation. This enables efficient deployment of models in resource-constrained environments and facilitates faster inference.
The Model agent also conducts model evaluation to assess the performance and generalization of the trained models. Various evaluation metrics and techniques are employed to ensure the models meet the user-desired criteria and deliver reliable predictions.
Furthermore, the Model agent facilitates model visualization, providing visual representations and summaries of the models' architecture, learned representations, and decision boundaries. This aids in model interpretation and understanding, allowing users to gain insights into the model's behavior. 


{\noindent \bf Server Agent}

The Server Agent handles the deployment of models based on user-defined online service requirements. 
Similar to the Data and Model agents, the Server agent receives requirements from the Task agent and autonomously performs planning and actions. 
Specifically, the Server agent conducts resource estimation, dynamically assessing the computational and memory resources required for model deployment. This estimation considers factors such as server specifications and expected service concurrency. By accurately estimating resource needs, the Server agent ensures efficient utilization of available infrastructure and prevents resource bottlenecks during model serving. 
Furthermore, the Server agent is responsible for model conversion, ensuring compatibility and efficiency during the deployment process. It performs conversions from frameworks like PyTorch or TensorFlow to formats such as ONNX and TensorRT. This enables seamless integration with different runtime environments and optimizes model inference performance. 
Moreover, the Server agent focuses on interface document preparation to facilitate collaboration between engineering and business teams. It prepares comprehensive and parameterized service invocation interfaces, enabling seamless communication and integration of the deployed models into various applications and systems. These interface documents serve as a reference for both technical implementation and business integration.
In summary, the Server agent ensures efficient resource allocation, seamless deployment, and effective integration of the models into real-world applications. Through its contributions, the Server agent strengthens the practicality and usability of the TrainerAgent system. 

\section{Experiments}
To validate the effectiveness of our TrainerAgent, we conducted experiments on real-world business scenarios from Taobao, a popular e-commerce platform, in both computer vision (CV) and natural language processing (NLP) domains. Specifically, we focused on classical discriminative and generative tasks including Visual Grounding, Image Generation, and Text Classification. Additionally, we also tested the system's ability to handle challenging tasks that could lead to failure. 
In our experiments, we utilize GPT-4 as a standalone agent within the TrainerAgent system. Each agent is individually configured with a profile, also known as a system prompt. Users directly interact with the TrainerAgent system through dialogue, ultimately completing the model training process. 
Note that although our experiments were conducted specifically within the Taobao, the TrainerAgent system can be generalized and applied in various real-world scenarios. 

\begin{figure*}[t]
\begin{center}
\includegraphics[width=2.1\columnwidth]{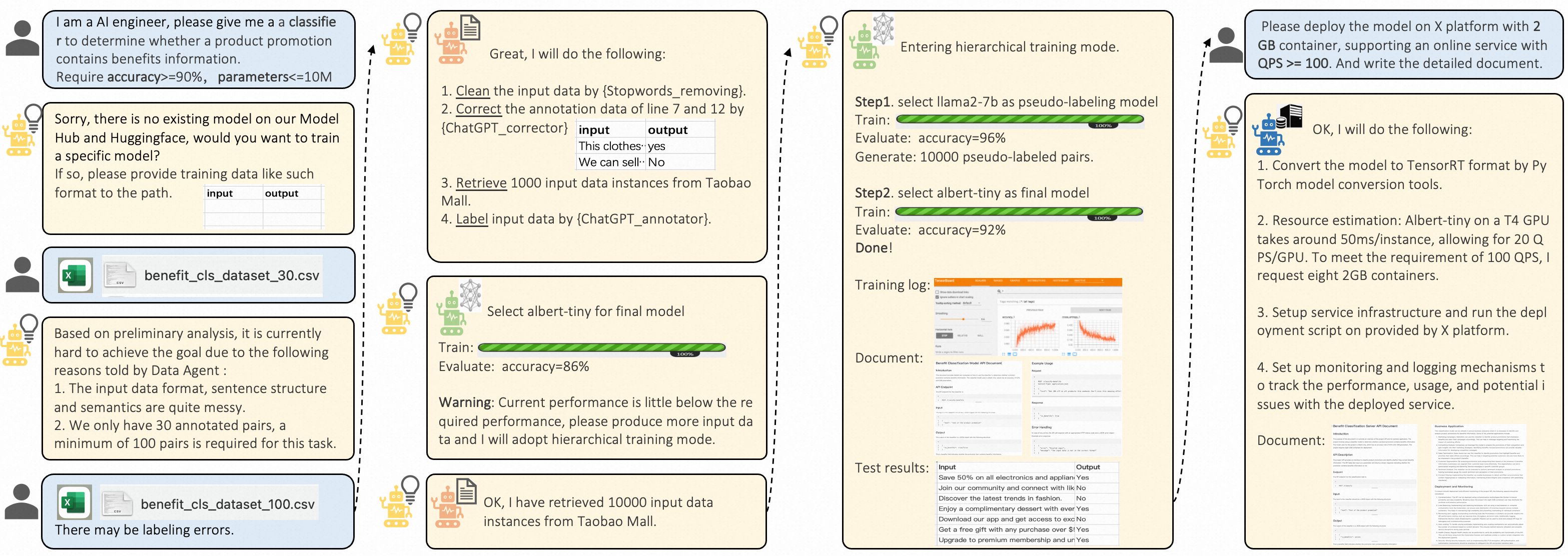}
\end{center}
   \caption{Interaction with TrainerAgent in Text Classification Task.}
\label{fig:text}
\end{figure*}

\subsection{Visual Grounding}

Visual Grounding (VG) \cite{RohrbachECCV16Grounding,Chen2017Query,Hu2017CMN,Yu2018MAttNet,Karpathy17Deep,YangICCV19FAOA,Liu21Relation_weak,Deng2021TransVG,zhaotowards} aims to localize the objects on an image according to a text query. Similarly, Product Grounding \cite{li2023date} aims to ground product, internally constructed within Taobao previously, which is more simple than a completely new task. Thus, we input all requirement into the system for both the training and deployment processesy to test the capabilities of TrainerAgent.

As shown in Figure \ref{fig:vg}, the system successfully accomplishes the internally constructed Product Grounding task, addressing the specific performance and deployment requirements presented by the user. This highlights the system's capability to handle task specifications and deliver satisfactory results. TrainerAgent system exemplifies a collaborative, adaptive, and efficient multi-agent framework for AI model development, embodying advantages in task analysis, data processing, model training, and server deployment. Each Agent is designed to perform specialized tasks, collaborating and communicating with each other to make optimal decisions collectively. 
Specifically, the Task Agent exhibits effective preliminary planning by parsing the task requirements and initiating the planning process. Furthermore, the Task Agent demonstrates great interaction and collaboration with the other four specialized agents. This emphasizes the system's ability to facilitate coordination and communication among different agents, ensuring a smooth workflow and efficient task execution. In addition, the other three specialized agents (Data Agent, Model Agent, and Server Agent) each perform their designated roles in a competent manner. The Data Agent retrieves relevant product grounding dataset from internal databases and enhances it through image and text preprocessing techniques. The Model Agent selects a pre-trained model from an internal library, trains and evaluates it against the specified criteria. The Server Agent undertakes various tasks such as model format conversion, resource estimation, service infrastructure setup, API documentation writing, and continuous monitoring. This highlights the system's capability to delegate specific responsibilities to the specialized agents, ensuring that each agent contributes to the overall success of the task.

The qualitative analysis of the Visual Grounding task in the proposed TrainerAgent system demonstrates its ability to effectively handle internally constructed tasks, perform preliminary planning, and facilitate collaboration among different agents. The specialized agents also showcase their competence in fulfilling their assigned responsibilities. These features collectively contribute to the overall functionality and effectiveness of the TrainerAgent system.

In addition, we conducted experiments on Image Generation, which are presented in the Appendix.

\subsection{Text Classification}
In this part, we will explore the pure NLP domain, where ChatGPT's powerful capabilities make handling NLP tasks more convenient, requiring less reliance on external tools compared with vision or audio domain. For instance, ChatGPT can directly analyze textual data and perform tasks such as data generation, augmentation, and error correction. In the following, we take the example of a classic text classification task to illustrate how TrainerAgent deals with the scarcity of annotated data, as shown in Figure \ref{fig:text}. 

In this experiment, we utilize the TrainerAgent to develop a classifier for determining whether a product promotion contains benefits information. Unlike the scenario where the user provides requirements all at once in Visual Grounding, this experiment is conducted in a step-by-step interactive manner involving more human participation, with the system adapting to the user's requirements and providing assistance throughout the process.
The User initially expresses their need for a classifier with an accuracy of at least 90\% and a parameter count below 10 million. The Task Agent performs an initial task analysis and conducts preliminary model and data searches. However, no existing model is found that meets the user's requirements. Instead of providing an unsatisfactory solution, the Task Agent suggests training a specific model using the available data.
The Data Agent plays a crucial role in this experiment. It assists the Task Agent in analyzing the data and determines that the input data format, sentence structure, and semantics are messy. Additionally, the Data Agent identifies that the initial dataset of 30 labeled pairs is insufficient for training an accurate model. Based on past experience and data quality assessment, the Data Agent recommends a minimum of 100 labeled pairs for the task.
The User responds by providing an updated dataset of 100 labeled pairs, acknowledging that there might be labeling errors present. The Data Agent proceeds to improve the data quality by performing several tasks. Firstly, it cleans the input data by removing stopwords to enhance the model's performance. Secondly, the annotation data of lines 7 and 12 are corrected using the internal ChatGPT\_corrector tool, ensuring accurate labeling. Thirdly, to expand the dataset, the Data Agent retrieves an additional 1000 input data instances from Taobao Mall. Lastly, the input data is automatically labeled using the internal ChatGPT\_annotator tool. 
The Model Agent, responsible for model selection and training, makes a decision based on the user's requirement for a small parameter count. It chooses the albert-tiny model for training. However, during the evaluation phase, the model's accuracy is found to be 86\%, falling short of the desired 90\% accuracy. To address this issue, the Model Agent autonomously selects a hierarchical training mode, optimizing the training process for the final small model. In this mode, the llama2-7b model is employed for pseudo-labeling, generating a larger labeled dataset. Subsequently, the albert-tiny model is trained on this expanded dataset. The final evaluation yields an accuracy of 92\%, meeting the user's requirement.
During the experiment, the User makes an additional request to deploy the trained model on a specific platform with a 2GB container. The Server Agent swiftly responds by converting the model to TensorRT format using PyTorch model conversion tools. Resource estimation determines that to achieve a minimum QPS of 100, eight 2GB containers are required. The Server Agent sets up the service infrastructure, executes the deployment script provided by the platform, and implements monitoring and logging mechanisms to track the deployed service's performance, usage, and potential issues.
This experiment demonstrates the effectiveness of the TrainerAgent system in developing a text classifier. The iterative and interactive nature of the experiment allows for a smoother and more user-involved process compared to a one-time requirement submission. The Task Agent's analysis, the Data Agent's data-related tasks, and the Model Agent's autonomous training mode selection showcase the system's capabilities and adaptability. Additionally, the system effortlessly accommodates the User's request for deployment, demonstrating the ease of integrating sudden deployment requirements into the system's workflow. 
In addition to the experiments shown above, our system can be applied to many multimodal tasks \cite{yin2021simulslt,huang2022prodiff,huanggenerspeech,jin2020dual,lin2021simullr,xia2022video }.

\subsection{Failed or Refused Tasks}

In this part, we will introduce tasks that our systems might fail or refuse to do. 
Our system may fail to solve pretty challenging task. Suppose a user requests a tough task (e.g. Video Question Answering \cite{yang2003videoqa}), however, there is no labeled data available for training the model, and the user demands a high accuracy for the task. After conducting an extensive analysis, our Task Agent can autonomously determine that it cannot meet the user's requirements due to the lack of labeled data and the performance limitations of existing models.
Despite conducting extensive data and model searches, the Agents are unable to find suitable resources to meet the user's requirements. To overcome this limitation, the Agents request user intervention, such as manually annotating more data to improve model performance. If the user does not provide the necessary assistance, our system will appropriately conclude that it cannot fulfill the task due to the lack of available resources and training data.
Additionally, our TrainerAgent will refuse to implement tasks for ethical reasons. In order to uphold ethical standards and ensure the safety of users, our system will refuse to perform certain tasks. For example, if a user requests the system to generate content that is harmful, offensive, or violates ethical norms, the Task Agent understands the request and its potential consequences. The Agent recognizes the importance of responsible AI usage and the potential harm that such generated content can cause. It prioritizes user well-being and the ethical implications of the task. Therefore, the Agent firmly refuses to comply with the request, ensuring that the system does not contribute to the dissemination of harmful or inappropriate content. The Agent emphasizes the ethical guidelines and ethical responsibility of the system, fostering a safe and supportive environment for users. 

By incorporating the Agent's understanding and decision-making process, these detailed explanations showcase how the system assesses tasks, recognizes limitations, and considers ethical implications. This enhances the system's user-centric approach and responsible deployment of AI models.

\section{Conclusion}
In this paper, we present a pioneering TrainerAgent system that revolutionizes the process of AI model development. This system leverages a multi-agent framework comprising Task, Data, Model, and Server agents, each playing a pivotal role in streamlining the development process. By comprehensively analyzing user-defined tasks, data, and requirements, our TrainerAgent optimizes models from both data and model perspectives, resulting in the creation of highly satisfactory models that can be seamlessly deployed as online services. The proposed TrainerAgent system offers a plethora of advantages over traditional model development approaches. Firstly, it dramatically reduces the time and effort required to develop customized models, opening the doors to AI for non-experts and accelerating the pace of innovation. Secondly, it ensures that the produced models meet the desired criteria, such as accuracy and speed, through a comprehensive optimization process. This not only boosts the quality and effectiveness of the models but also enhances the overall user experience. However, our system still has several limitations.

Lower Success Rate: Currently, our TrainerAgent system relies on pre-established local model running scripts, which limits its ability to successfully run on any open-source code available on platforms like GitHub. To address this limitation, we are committed to enhancing the system's capability to automatically understand documentation, such as readme files, and autonomously execute the code, thereby improving the success rate of model implementation.

Dependence on Human Interaction: The TrainerAgent system still requires interaction with humans to ensure optimal performance and customization. However, as the system undergoes iterative improvements, we aim to minimize this dependence and ultimately achieve an end-to-end model training and deployment process. By doing so, we will reduce the need for extensive manual intervention, enhancing the system's autonomy and usability.

Limited Generalization: While our system demonstrates effectiveness in various tasks, its generalization across a wide range of domains and applications may be limited. The current version of TrainerAgent focuses on discriminative and generative tasks in computer vision and natural language processing. To address this limitation, future iterations of the system will incorporate additional domains and expand the scope of task applicability, allowing for more diverse and comprehensive model development.

Ethical Implications: As with any AI system, our TrainerAgent system raises ethical considerations. While efforts are made to ensure the system adheres to ethical guidelines, there is always a possibility of unintended consequences or biases in decision-making. We are committed to ongoing research and development to address these ethical implications and incorporate safeguards to mitigate potential risks.

Despite these limitations, our TrainerAgent system represents a significant step forward in customizable and efficient model training. Through continuous improvements and addressing these limitations, we aim to enhance the system's performance, adaptability, and overall impact in both academic and industry settings.


\clearpage

{\small
\bibliographystyle{ieee_fullname}
\bibliography{main.bib}
}

\end{document}